\documentclass[dvipsnames]{article}
\usepackage{arxiv}

\usepackage{amsmath,amssymb,amsthm,amsfonts,mathtools}
\usepackage{graphicx}
\usepackage{textcomp}
\usepackage{braket}
\usepackage{booktabs}
\usepackage{multirow}
\usepackage{xspace}

\usepackage{bm}

\DeclareMathAlphabet{\mathdutchcal}{U}{dutchcal}{m}{n}
\SetMathAlphabet{\mathdutchcal}{bold}{U}{dutchcal}{b}{n}
\DeclareMathAlphabet{\mathdutchbcal}{U}{dutchcal}{b}{n}
\DeclareMathAlphabet{\mathpzc}{OT1}{pzc}{m}{it}

\makeatletter
\let\MYcaption\@makecaption
\makeatother
\usepackage{subcaption} %
\makeatletter
\let\@makecaption\MYcaption
\makeatother

\usepackage[
    backend=biber
  ,giveninits=true                  %
  ,url=false, isbn=false, doi=false %
  ,maxnames=99                       %
  ,minnames=3                       %
  ,sorting=none             %
  ,date=year                %
  ,style=ieee
  ]{biblatex}
\AtEveryBibitem{                      %
  \iffieldequalstr{eprinttype}{jstor}
  {\clearfield{eprint}}
  {}
  \clearfield{urlyear}
  \clearfield{urlmonth}
  \clearfield{url}
}
\renewrobustcmd*{\bibinitdelim}{\,} %
\usepackage{lipsum}

\usepackage[boxed,ruled,vlined,linesnumbered]{algorithm2e}
\DontPrintSemicolon
\SetKwProg{Fn}{function}{}{}
\SetKwFunction{FnSampleFree}{SampleFree}
\SetKwFunction{FnRestartArm}{RestartArm}
\SetKwFunction{FnPickArm}{PickArm}
\SetKwFunction{FnRewire}{Rewire}
\SetKwFunction{FnNearest}{Nearest}
\SetKwFunction{FnRestartArm}{RestartArm}
\SetKwComment{Comment}{$\triangleright$\ }{}
\SetKwInput{KwInit}{Initialise}

\usepackage{cleveref}                                        %
\crefname{assumption}{assumption}{assumptions}
\crefname{problem}{problem}{problems}
\crefname{algorithm}{Alg.}{Algs.}
\Crefname{algorithm}{Algorithm}{Algorithms}
\crefname{figure}{Fig.}{Figs.} %
\crefformat{equation}{(#2#1#3)}
\crefrangeformat{equation}{(#3#1#4) to~(#5#2#6)}
\crefmultiformat{equation}{(#2#1#3)}%
{ and~(#2#1#3)}{, (#2#1#3)}{ and~(#2#1#3)}

\usepackage{microtype}

\usepackage{tkz-graph}
\usetikzlibrary{shapes.multipart, positioning, decorations.markings, arrows.meta, calc}

\microtypesetup{activate={true,nocompatibility},final,tracking=true,kerning=true,factor=1100,stretch=10,shrink=10}
\usepackage{etoolbox}
\makeatletter
\pretocmd{\NAT@citexnum}{\@ifnum{\NAT@ctype>\z@}{\let\NAT@hyper@\relax}{}}{}{}
\makeatother

\newcommand*\given[1][]{\:#1\vert\:}
\newcommand*\prob[1][]{\mathbb{P}}
\newcommand*\norm[1]{\big\lVert#1\big\rVert}
\DeclareMathOperator*{\argmax}{\arg\max}
\DeclareMathOperator*{\argmin}{\arg\min}

\newcommand*\State{\mathbf{X}}
\newcommand*\Landmark{\mathbf{L}}
\newcommand*\landmark{\mathbf{l}}
\newcommand*\Obs{\mathbf{Z}}
\newcommand*\obs{\mathbf{z}}

\begin{document}

\newcommand{\shortheadtitle}{
	A Review on Visual-SLAM: Advancements in Semantic Understanding
}

\title{
	A Review on Visual-SLAM: \\Advancements from Geometric Modelling to Learning-based Semantic Scene Understanding
}

\author{
	Tin Lai\\
	School of Computer Science\\
	University of Sydney\\
	Australia \\
}

\maketitle

\begin{abstract}
	Simultaneous Localisation and Mapping (SLAM) is one of the fundamental problems in autonomous mobile robots where a robot needs to reconstruct a previously unseen environment while simultaneously localising itself with respect to the map.
	In particular, Visual-SLAM uses various sensors from the mobile robot for collecting and sensing a representation of the map.
	Traditionally, geometric model-based techniques were used to tackle the SLAM problem, which tends to be error-prone under challenging environments.
	Recent advancements in computer vision, such as deep learning techniques, have provided a data-driven approach to tackle the Visual-SLAM problem.
	This review summarises recent advancements in the Visual-SLAM domain using various learning-based methods.
	We begin by providing a concise overview of the geometric model-based approaches, followed by technical reviews on the current paradigms in SLAM.
	Then, we present the various learning-based approaches to collecting sensory inputs from mobile robots and performing scene understanding.
	The current paradigms in deep-learning-based semantic understanding are discussed and placed under the context of Visual-SLAM.
	Finally, we discuss challenges and further opportunities in the direction of learning-based approaches in Visual-SLAM.
\end{abstract}

\section{Introduction}

Autonomous navigation in mobile robots has become an active research field in recent years due to the advancements in material science for robot construction, compact battery size for more prolonged duration remote operations, and the increases in computational hardware for powering algorithmic and artificial intelligent methods.
Mobile robots are capable of navigating within some environments to perform their respective objectives.
Most autonomous robots need to move around in the environment while respecting the environment.
Robots ``{see and understand}'' the world through collecting information from the attached sensors and making sense of the readings.
Specifically, for robots to interact with the real world, they need some capabilities of understanding the scene geometrically and interpreting it semantically.
Accurate localisation of the robot is essential for the robot to compute suitable actions, and having a good knowledge or perception of the environment allows the robot to react to its surroundings.

Mobile robots typically receive sensor information from their attached sensors~\cite{borenstein1997_MobiRobo}, for example, in the form of 2D projections of image frames or 3D spatial points from high-frequency LiDAR scans~\cite{kolhatkar2021_ReviSLAM}.
However, this perceived information is often insufficient for the robot to navigate as it lacks the geometric understanding and reconstruction of the scene.
Geometric modelling is especially essential in complex tasks where the robot needs to localise itself with respect to the modelled map of the scene to navigate and accomplish its objectives.
For example, mobile robot navigation often requires the robot to maintain a map representation for performing planning in tasks such as motion planning~\cite{lai2021_AdapExpl}, navigating mobile robots~\cite{garrido2006_PathPlan}, moving robotics arms~\cite{lai2018_BalaGlob}, or even autonomous cars~\cite{katrakazas2015_RealMoti}.
It is infeasible for the robot to perform safe autonomous operations in a dynamic environment if it lacks the ability to perceive and make sense of the potential obstacles.
This is especially important for the robot to operate in a novel region without any prior information about the environment, for example, in planetary exploration or search and rescue operations.

Difficulty in the geometric reconstruction of the environment while navigating within the unknown environment often arises due to issues like sensor modalities, misalignment of map representations, or observation noises~\cite{flint2010_GrowSema, lothe2009_GeogRefe, weingarten2005_EKFb3D}.
Simultaneous Localisation and Mapping (SLAM) addresses the problem of an online incremental process where the mobile robot needs to refine the reconstructed map and its current location iteratively by observing more of the unknown environment using various sensors~\cite{2015_SensTech}.
In Visual-SLAM, the most straightforward representation of the environment often consists of a collection of sparse 3D points processed continuously during the navigation.
A robotic system collects visual inputs for Visual-SLAM while operating in some unknown environment~\cite{hong2021_VisuSLAM}, which constructs a map of the surroundings using various sensors while simultaneously estimating its position concerning the environment.
The constructed map can be used for mapping a novel environment or for the robot to plan for its mission with autonomy.
Such a system can maintain stability, plan for its own movements, and reacts to dynamic changes in the surroundings without human intervention~\cite{bavle2021_SLAMSitu}.

SLAM Mobile robot semantic scene involves both the domain of robotics, computer vision and sensor technologies.
A wide range of sensors can be used for SLAM.
For example, most autonomous vehicles use Light Detection and Ranging (LiDAR) sensor~\cite{hess2016_RealLoop} or stereo cameras~\cite{engel2015_LargDire} to perceive the surrounding environment during navigation.
LiDAR can often provide a more accurate environment representation by providing a 3D point cloud with ranging measurements.
Traditional LiDAR sensors are rarely used in consumer-grade mobile robots due to the high cost; however, the advancements in manufacturing have enabled LiDAR sensors to become more common in mobile robots navegation~\cite{pang2019_LowcHigh, zhu2021_CamvLowc}
In contrast, stereo cameras are more ubiquitous as the manufacturing cost is much lower than LiDAR.
On the other hand, the hybridisation of multiple sensors is shown to enhance the localisation, and mapping performance in most SLAM approaches significantly~\cite{xiong2016_HybrLida}.
Hybridisation visual sensors with classical proprioceptive sensors such as IMU or odometers can often reduce the localisation drift due to the cumulative error of these relative positioning approaches.
Rather than having some superior type of sensors, different sensing methods have their relative strength and weaknesses~\cite{moleski2020_TrilPosi, su2017_GlobLoca}.
For example, laser scanners in LiDAR are efficient for obstacle detection but are highly sensitive to weather conditions like rain.
In contrast, RGB cameras can extract semantic meaning from the captured images but are sensitive to lighting conditions.
Therefore, most sensors are complementary, and it is an open research question of matching and balancing each type of sensor with their respective strengths and weaknesses.

Semantic scene understanding is neglected in traditional approaches, which only focus on the geometric reconstruction of the environment.
Rather than treating the collecting points that carry no relationship, semantic understanding assigns higher-level meanings to the collected data~\cite{yang2016_PopuSlam}.
Real-world environments often contain many structures and objects that carry high-level semantic information that is helpful to act as landmarks in SLAM.
Assignment semantic meaning can be helpful in both reconstructing the scene by inferring missing information and providing complementary information for reconstructing the scene~\cite{gonzalez2021_S3laStru, liao2022_SOSLSema}.
Moreover, reconstructing the geometric representation of the scene and their respective semantic meaning can be helpful for the mobile robot to make higher-level decisions on selecting landmarks that are suitable for the environment and informing the robot planner on deciding its mission route.

This review article provides an overview of the current state-of-the-art Visual-SLAM paradigms and models with a focus on sensor fusion.
This paper is organised as follows.
We begin by first providing a concise summary of the theory behind the SLAM process and the formulation of the geometric modelling of the surrounding environment.
Then, we present an overview of the evolving SLAM paradigms throughout recent years, including both approaches in pure geometric reconstruction and semantic scene understanding using a deep-learning model.
Finally, we discuss the current state-of-the-art Visual-SLAM models to understand our progress and future direction in SLAM.

\section{Simultaneous Localisation and Mapping}

\emph{Simultaneous Localisation and Mapping} (SLAM) is a problem where a robot needs to operate in an unknown environment to construct a map while estimating its uncertain location \cite{feder1999_AdapMobi}.
SLAM is a fundamental problem in numerous robotics applications that needs the robot to autonomously navigate within some environment and interact with the real world.
In the following, we will introduce the problem setup in SLAM, followed by a formalisation of the fundamental theory behind SLAM algorithms.

\subsection{Problem Setup}

The problem's difficulty comes from the recursive dependency: constructing a map often depends on the robot observing the environment from some \emph{known location}, while state estimation also often requires a robot to infer its location by relying on some \emph{known landmarks}.
SLAM algorithm estimates the sensor motion and simultaneously reconstructs the geometrical structure of the visited area.
\citeauthor{chatila1985_PosiRefe}~\cite{chatila1985_PosiRefe} first formularised the problem setup in~\citeyear{chatila1985_PosiRefe} for mobile robots navigation.
The problem lies in the need to model the environment and locate itself correctly through the inaccuracies introduced by the sensors.
The proposed methodology defines a general principle to deal with uncertainties in the collected data and for a mobile robot to define its reference landmarks while exploring the environment.

The fundamental idea of SLAM lies in using landmark correlations, data association, and loop-closure to reduce the uncertainties about its previously visited area and poses~\cite{frese2006_DiscSimu}.
Traditional techniques for sequential state estimation include Kalman filter~\cite{welch1995_IntrKalm}.
Kalman filter is an optimal state estimation technique in a linear system with Gaussian noise.
Practical implementations of SLAM often use the extended Kalman filter (EKF) for state estimation \cite{ribeiro2004_KalmExte}, which is advantageous because the Gaussian assumption allows EKF to be measured analytically.
If the system has non-Gaussian noise, the Kalman filter is still the optimal \emph{linear} filter but performs worse than other techniques.
For nonlinear systems, methods such as particle filter~\cite{carpenter1999_ImprPart} can be a more flexible alternative as it does not rely on any local linearisation technique or crude functional approximation.
However, the higher performance comes with a higher computational effort than Kalman filters.
In particle filtering, we need to perform weighted sampling to estimate the distribution of the robot state rather than having an analytical solution to obtain the robot state distribution by using the mean and covariance matrix in a Gaussian distribution.

There are multiple metrics to measure the benefit of actions.
For example, \emph{A-optimality} measures the trace of the covariance matrix~\cite{sim2005_GlobAopt}, which is equivalent to minimising the mean squared error between the data and model parameters.
\emph{D-optimality}, on the other hand, minimises the determinant of the covariance matrix~\cite{bryson2008_ObseAnal}, which is equivalent to minimising the entropy of the SLAM system~\cite{carrillo2012_CompUnce}.
For example, we can utilise the building structure lines as features for localisation and mapping, which can encode the global orientation information constrains the robot's heading over time.
These features help eliminate the accumulated orientation errors and reduce the position drift in SLAM algorithms.
In SLAM, the concept of \emph{loop closure} can also reduce drift errors by allowing the robot to reset its estimated state by revisiting a known portion of the map~\cite{bryson2008_ObseAnal}.
Active SLAM methods often exploit this property by guiding the robot to regions that allow the robot to close the loop~\cite{lenac2016_FastActi}, which can significantly reduce the localisation error~\cite{stachniss2004_ExplActi}.
Autonomous navigation in an indoor environment often requires multiple sensory inputs and actuating outputs.
Wheeled ground mobile robots are often designed with a differential drive base that uses DC geared or stepper motors for their driving wheels.
Mobile robots collect data from onboard sensors like wheel encoders, initial measurement units (IMU), RGB cameras for visual inputs, or LiDAR as remote sensing to measure ranges.

Loop-closing can detect if a given keyframe had been seen previously~\cite{cummins2008_FABMProb,mei2009_ConsEffi}.
Loop Closure can be formulated as an optimisation problem, such as a nonlinear least squares problem that matches the current scans with previously visited areas.
One reason that loop closing is hard in SLAM is that the internal estimates can, despite best efforts, be in gross error.
Loop closing is essentially a data association problem where a positive loop closure occurs when the robot \emph{recognises} the local scene to be one that it has previously visited.
Traditional feature-based SLAM uses simple geometric primitives such as corners or lines as features.
When a loop closure is detected, it acts as an opportunity to constrain the robot's internal estimate of its current state with respect to the map.

\begin{figure}
	\centering
	\begin{tikzpicture}[
			mypath/.style={postaction=decorate},
			state/.style={circle, draw, minimum size=0.5cm, fill=#1, anchor=center},
			state/.default=blue!30,
			conn/.style={circle, draw, scale=0.5, anchor=center, fill=black},
			font=\sffamily
		]
		\node[conn, label={$\prob_0$}] (p0) {};
		\node[state, right=of p0, label={$\State^w_1$}] (t1) {};
		\node[conn, right=of t1, label={$\mathbf{u}_1$}] (u1) {};
		\node[state, right=of u1, label={$\State^w_2$}] (t2) {};
		\node[right=of t2, label={}] (dots) {$\ldots$};
		\node[state, right=of dots, label={$\State^w_{N-1}$}] (tNminus1) {};
		\node[conn, right=of tNminus1, label={$\mathbf{u}_{N-1}$}] (uNminums1) {};
		\node[state, right=of uNminums1, label={$\State^w_{N}$}] (tN) {};
		\node[state=green!30, below right=of t1, xshift=.4cm, yshift=-.7cm, label={$\landmark_{1}$}] (l1) {};
		\node[conn, label={[anchor=east]$\obs_{1}$}] (z1) at ($(t1)!0.5!(l1)$)  {};
		\node[conn, label={[anchor=west]$\obs_{2}$}] (z2) at ($(t2)!0.5!(l1)$)  {};
		\node[state=green!30, below right=of dots, xshift=-.9cm, yshift=-.7cm, label={$\landmark_{2}$}] (l2) {};
		\node[conn, label={[anchor=east]$\obs_{3}$}] (z3) at ($(t2)!0.5!(l2)$)  {};
		\node[conn, label={[anchor=west]$\obs_{4}$}] (z4) at ($(tNminus1)!0.5!(l2)$)  {};
		\foreach \i/\j in {
				p0/t1, t1/u1, u1/t2, t2/dots, dots/tNminus1, tNminus1/uNminums1, uNminums1/tN,%
				t1/z1, z1/l1, l1/z2, z2/t2, t2/l2, l2/z4, z4/tNminus1%
			}
		\draw[mypath] (\i) -- (\j);
	\end{tikzpicture}
	\caption{
		Formulating the visual-SLAM problem with a factor graph,
		where the camera poses are denoted as $\State^w_i$ and landmarks as $\landmark_j$.
		The observations of the landmarks and odometry at various camera poses are denoted as $\obs_k$ and $\mathbf{u}_i$, respectively.
		The prior belief on the initial pose is denoted as $\prob_0$, and the joint probability distribution of the MAP problem can be computed to the product of the depicted factors.
		\label{fig:factor-graph}}
\end{figure}
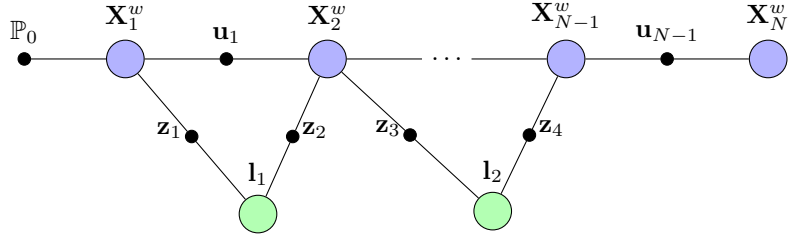

\subsection{SLAM Formulation}

SLAM is a multi-discipline problem that spans both the computer vision and robotics domain and is traditionally formulated as a \emph{maximum a posterior} (MAP).
In Visual SLAM, we define $\State = \{\State^w_i\}^N_{i=1}$ as the trajectory of the robot over time, where $\State^w_i$ denote the pose of the robot parameterised in the set of rigid Euclidean transformations $\mathbb{SE}(3)$.
Let $\Landmark=\{\mathbf{l_j}\}^M_{j=1}$ denote the set of landmarks parameterised by their appropriate representation space, $\Obs=\{\obs_k\}^K_{k=1}$ be the set of observations of the detected landmarks, and $\mathbf{U}=\{u\}_{i=1}^{N-1}$ be the set of odometry measurements between robot poses.
The observations $\Obs$ of the landmarks are collected under some observation model $h_k(\cdot)$, given by
\begin{equation}
	\obs = h_k(\State_{i_k}, \landmark_{j_k}) + \epsilon_k
\end{equation}
where $\State_{i_k}$ and $\landmark_{j_k}$ denote the actual robot state and landmark pose, and $\epsilon_k$ is a random measurement noise.
The solution to the SLAM problem is the optimal MAP estimation of
\begin{equation}\label{eq:map}
	\State^*, \Landmark^*
	= \argmax_{\State, \Landmark}
	\prob(\State, \Landmark \given \Obs, \mathbf{U})
\end{equation}
where $\prob(\State, \Landmark \given \Obs, \mathbf{U})$ is the joint probability of all latent estimate variables given all of our previous observations and measurements.
For a classical SLAM problem without odometry measurments~\cite{dellaert2017_FactGrap}, we can rewire~\cref{eq:map} as
\begin{align}
	\State^*, \Landmark^*
	 & = \argmax_{\State, \Landmark}
	\prob(\State, \Landmark \given \Obs) \\
	 & = \argmax_{\State, \Landmark}
	\prob(\Obs \given \State, \Landmark)
	\prob(\State, \Landmark) \label{eq:map-with-likelihood}
\end{align}
where $\prob(\Obs \given \State, \Landmark)$ is the likelihood of the obtaining the measurement $\Obs$ given $\State$ and $\Landmark$, and $\prob(\State, \Landmark)$ being the prior knowledge on $\State$ and $\Landmark$.
Assuming that each observation $\obs_k$ is independent, we can then compute~\cref{eq:map-with-likelihood} as
\begin{align}\label{eq:map-factor-graph-with-likelihood}
	\State^*, \Landmark^*
	 & = \argmax_{\State, \Landmark}
	\prod^K_{k=1} \prob(\obs_k \given \State_k, \Landmark_k)
	\prob(\State, \Landmark).
\end{align}

\subsection{Factor Graph and Loop-Closure}

\emph{Factor graph}~\cite{kschischang2001_FactGrap} represents an essential part of modern approaches to address the probabilistic SLAM problem by factorisation of and inference over arbitrary distribution functions.
A factor graph $\mathcal{G}(\mathcal{V}, \mathcal{F}; \mathcal{E})$ is a bipartite graph that determines the factorisation of variables from a global function into product of local functions.
Specifically, the set of vertices $\mathcal{V}$ in the graph $\mathcal{G}$ represents the latent variables that participate in the estimation process.
The set of factors $\mathcal{F}$ represents the prior knowledge regarding variable nodes and constraints between nodes, where the connections between nodes are represented by the set of edges $\mathcal{E}$.

We can represent a classical SLAM problem as a factor graph as depicted in~\cref{fig:factor-graph}, where the joint probability distribution of the MAP estimation is factorised as a product over observation factors.
Using the factor graph notation, we can rewrite the MAP formulation in~\cref{eq:map} as
\begin{align}\label{eq:map-factor-graph}
	\State^*, \Landmark^*
	 & = \argmax_{\State, \Landmark}
	\prob(\State, \Landmark \given \Obs) \\
	 & = \argmin_{\State, \Landmark}
	\sum^K_{k=1} \norm{ h_k (\State^w_{i_k}, \landmark_{j_k}) \ominus \obs_k}^2_{\mathbf{\Sigma}_k}
\end{align}
where $h_k$ denote the $k$\textsuperscript{th} factor of observing a landmark $\landmark_{j}$ from the camera pose $\State^w_{i}$ with the sensor model $\obs_k$, the notation $\norm{\cdot}^2_{\mathbf{\Sigma}}$ denote the squared Malnalanobis norm with covariance matrix $\mathbf{\Sigma}$, and $\ominus$ is the difference operator in the target measurement space.

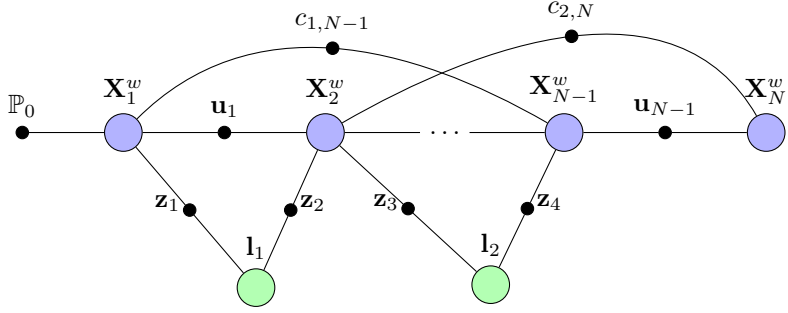
\begin{figure}
	\centering
	\begin{tikzpicture}[
			mypath/.style={postaction=decorate},
			state/.style={circle, draw, minimum size=0.5cm, fill=#1, anchor=center},
			state/.default=blue!30,
			conn/.style={circle, draw, scale=0.5, anchor=center, fill=black},
			font=\sffamily
		]
		\node[conn, label={$\prob_0$}] (p0) {};
		\node[state, right=of p0, label={$\State^w_1$}] (t1) {};
		\node[conn, right=of t1, label={$\mathbf{u}_1$}] (u1) {};
		\node[state, right=of u1, label={$\State^w_2$}] (t2) {};
		\node[right=of t2, label={}] (dots) {$\ldots$};
		\node[state, right=of dots, label={$\State^w_{N-1}$}] (tNminus1) {};
		\node[conn, right=of tNminus1, label={$\mathbf{u}_{N-1}$}] (uNminums1) {};
		\node[state, right=of uNminums1, label={$\State^w_{N}$}] (tN) {};
		\node[state=green!30, below right=of t1, xshift=.4cm, yshift=-.7cm, label={$\landmark_{1}$}] (l1) {};
		\node[conn, label={[anchor=east]$\obs_{1}$}] (z1) at ($(t1)!0.5!(l1)$)  {};
		\node[conn, label={[anchor=west]$\obs_{2}$}] (z2) at ($(t2)!0.5!(l1)$)  {};
		\node[state=green!30, below right=of dots, xshift=-.9cm, yshift=-.7cm, label={$\landmark_{2}$}] (l2) {};
		\node[conn, label={[anchor=east]$\obs_{3}$}] (z3) at ($(t2)!0.5!(l2)$)  {};
		\node[conn, label={[anchor=west]$\obs_{4}$}] (z4) at ($(tNminus1)!0.5!(l2)$)  {};
		\path (t1) edge [out=45,in=150] node[conn, label={$c_{1,N-1}$}] {} (tNminus1);
		\path (t2) edge [out=30,in=120] node[conn, label={$c_{2,N}$}] {} (tN);
		\foreach \i/\j in {
				p0/t1, t1/u1, u1/t2, t2/dots, dots/tNminus1, tNminus1/uNminums1, uNminums1/tN,%
				t1/z1, z1/l1, l1/z2, z2/t2, t2/l2, l2/z4, z4/tNminus1%
			}
		\draw[mypath] (\i) -- (\j);
	\end{tikzpicture}
	\caption{
	Visual-SLAM Bundle Adjustment (BA) in a factor graph.
	The potential odometry factor $\mathbf{u}_i$ constrain the relative camera poses with potential loop-closure factors $c_{i_1, i_2}$ where $i_1,i_2$ are the index of the camera poses.
	This figure demonstrate loop-closure factors $c_{1,N-1}$ between the camera pose $\State^w_1$ and $\State^w_{N-1}$, and $c_{2,N}$ between $\State^w_2$ and $\State^w_N$ for deciding whether the mobile robot had returned to a previously visited area.
	\label{fig:factor-graph-with-loop-closure}}
\end{figure}

The SLAM problem can be formulated as a Bayes net under the \emph{factor graph} formulation as factorisation and inference over probability distribution and functions~\cite{kaess2011_ISAMIncr,dellaert2017_FactGrap,folkesson2007_GrapSLAM,olson2006_FastIter, thrun2006_GrapSLAM}.
A \emph{factor graph} is a bipartite graph that characterises how a global multi-variable function can be factorised into a product of local functions.
Each blue and green node in~\cref{fig:factor-graph}, also known as \emph{variables}, represents the set of latent variables that need to be estimated, which in the case of SLAM are the state of the robots and the landmarks.
The node in-between the variables are known as \emph{factors}, which is the set of constraints and information between the variables.
We can use a factor graph to factorise a joint probability distribution over some random variables by encoding the inherent conditional independence of some local variables into the joint probability distribution.

The joint probability distribution of all the latent estimate variables of the SLAM problem can be written as
\begin{equation}\label{eq:joint-prob-dist-1}
	\prob(\State, \Landmark \given \Obs, \mathbf{U}) \propto
	\prob(\State^w_0)
	\prod^K_{k=1} \prob(\obs_k \given \State^w_{i_k}, \landmark_{j_k})
	\prod^N_{i=1} \prob(\State^w_{i} \given \State^w_{i-1}, \mathbf{u}_{i-1})
\end{equation}
where $\prob(\State^w_0) \equiv \prob_0$ is the prior belief on the robot's initial pose.
The $\prob(\obs_k \given \State^w_{i_k}, \landmark_{j_k})$ represents the effect of landmark observation $\obs_k$ given the data association $(i_k, j_k)$, and $\prob(\State^w_{i} \given \State^w_{i-1}, \mathbf{u}_{i-1})$ represents the state update given the motion model.
Assuming a zero-mean Gaussian observation noise for observation  $\Obs$ and odometry $\mathbf{U}$, we can rewrite~\cref{eq:joint-prob-dist-1} as
\begin{equation}
	\prob(\State, \Landmark \given \Obs, \mathbf{U}) \propto
	\underbrace{
	\prod^K_{k=1} \exp\left(-\frac{1}{2} \norm{ h_k (\State^w_{i_k}, \landmark_{j_k}) \ominus \obs_k}^2_{\mathbf{\Sigma}_k} \right)
	}_\text{effect of observations}
	\underbrace{
	\prod^N_{i=1} \exp\left(-\frac{1}{2} \norm{ f_o (\State^w_{i-1}, \mathbf{u}_{i-1}) \ominus \State^w_i}^2_{\mathbf{\Sigma}_o} \right)
	}_\text{effect of odometry}
\end{equation}
where $h_k$ is the sensor model, $f_o$is the motion model, $\mathbf{\Sigma}_k$ and $\mathbf{\Sigma}_o$ are the covariance matrix for the Gaussian noise in $\Obs$ and $\mathbf{U}$, respectively.

We can further factorise this joint probability distribution to obtain the optimal MAP estimation by solving the equivalent least-squares form of
\begin{align}
	\State^*, \Landmark^*
	 & = \argmax_{\State, \Landmark}
	\prob(\State, \Landmark \given \Obs, \mathbf{U})        \\
	 & = \argmin_{\State, \Landmark}
	- \log \prob(\State, \Landmark \given \Obs, \mathbf{U}) \\
	 & = \argmin_{\State, \Landmark}
	\underbrace{
	\sum^K_{k=1} \norm{ h_k (\State^w_{i_k}, \landmark_{j_k}) \ominus \obs_k}^2_{\mathbf{\Sigma}_k}
	}_\text{effect of observations} +
	\underbrace{
	\sum^N_{i=1} \norm{ f_o (\State^w_{i-1}, \mathbf{u}_{i-1}) \ominus \State^w_i}^2_{\mathbf{\Sigma}_o}
	}_\text{effect of odometry}
	,
\end{align}
which can be interpreted graphically as a factor graph as the one shown in~\cref{fig:factor-graph,fig:factor-graph-with-loop-closure}.

The depiction in~\cref{fig:factor-graph-with-loop-closure} indicates an instance of the classical bundle adjustment (BA)~\cite{hartley2003_MultView,trevor2014_OmniModu}.
In BA, the factor graph's variable nodes can be considered camera poses and 3D landmarks to minimise the re-projection error factors.
BA applications use sensor information from odometry in mobile robots or IMU to further improve the accuracy of the estimated robot trajectory.
In~\cref{fig:factor-graph-with-loop-closure}, the loop-closure factors can be extended to higher-level entities that impose some sophicated constraints and factors.
Loop-closure can often improve the consistency of the mapping results~\cite{stachniss2004_ExplActi} as they act as additional constraints during the factorisation of the joint distribution~\cite{sunderhauf2012_SwitCons}.

\section{Evolution of SLAM Techniques and Paradigms}

Various approaches in SLAM have been proposed throughout the years to address different challenges within the approach.
The following discusses the operational process of traditional SLAM algorithms and recent developments in SLAM paradigms.
A summary of their algorithmic approach, advantages and shortcoming are provided as follows.
\begin{itemize}
	\item
	      \textbf{Fast-SLAM} (\citeyear{montemerlo2002_FastFact})~\cite{montemerlo2002_FastFact} addresses the localisation problem by using a decomposing strategy for recursively estimating the full posterior distribution over robot pose and landmark location.
	      The algorithm performs exact factorisation of the posterior into a conditional landmark distribution and distribution over robot paths.
	      \emph{Advantage:} The complexity scales logarithmically with the number of landmarks on the map.
	      \emph{Disadvantage:} FastSLAM behaves like a non-optimal local search algorithm, where it is capable of producing consistent uncertainty estimates, but, in the long-term, it is unable to explore the state-space as a Bayesian estimator adequately.
	\item
	      \textbf{Extended Kalman Filter (EFK) SLAM} (\citeyear{paz2007_EKFSLAM})~\cite{paz2007_EKFSLAM} uses a divide and conquer approach to estimate a consistent state estimation by using state covariance to represent the real error in the estimation process.
	      EFK SLAM uses phase iterations of predictions, observation, and updates to perform state estimation in a Bayesian manner.
	      \emph{Advantage:} EFK SLAM often achieve a more consistent estimation than another approach as it computes the exact solution rather than using approximation, and the proposed approach tackles the combinatorial complexity.
	      \emph{Disadvantage:} Despite the consistent estimation, the approach uses a probabilistic inference approach for forecasting the current state, which might diverge from the actual current state.
	\item
	      \textbf{V-SLAM} (\citeyear{lategahn2011_VisuSLAM})~\cite{lategahn2011_VisuSLAM} computes a locally dense stereo correspondences from the potentially sparse raw representation.
	      The dense representation avoids the sparsity problem that often arises in operating SLAM with a sparse set of landmarks.
	      \emph{Advantage:} The computational overhead is relatively small, and the dense representation increases the robustness of the inner SLAM algorithm in a sparse environment.
	      \emph{Disadvantage:} The dense representation can be more sensitive to flaws or environmental changes to the environment.
	\item
	      \textbf{Large-Scale Direct (LSD) SLAM} (\citeyear{engel2015_LargDire})~\cite{engel2015_LargDire} aligns images directly with photoconsistency of high-contrast pixels.
	      LSD SLAM can concurrently estimate the depths at the pixels using static stereo and temporal multi-view stereo by utilising the camera motion.
	      \emph{Advantage:} The SLAM can operate directly at the pixel level rather than as a separate procedure for processing the captured images.
	      \emph{Disadvantage:} The procedure can be costly when computing the translational motion between frames.
	\item
	      \textbf{ORB-SLAM2} (\citeyear{mur-artal2017_OrbsOpen})~\cite{mur-artal2017_OrbsOpen} make use of multiple features from Monocular, Stereo and RGB-D cameras which greatly enhance the versatility of the method.
	      The algorithm uses bundle adjustment to create a 3D environment by extracting features from different images and placing them in 3D.
	      \emph{Advantage:} ORB-SLAM2 is highly versatile and can perform sensor fusion to improve detection quality.
	      The model includes loop-closure detection, keyframe selection and per-frame localisation, which enhance its robustness.
	      \emph{Disadvantage:} High processing cost, which might be costly for small systems.
	\item
	      \textbf{2D-LiDAR SLAM} (\citeyear{chan2018_Robu2D})~\cite{chan2018_Robu2D} is an algorithmic approach that uses a laser sensor to create a 2D view of its surroundings. The method uses laser and visual fusion to provide localisation by combining two kinds of laser-based SLAM and monocular camera-based SLAM.
	      \emph{Advantage:} The fusion allows high performance in spotting complex structures like hollow ceilings and can achieve high precision even at a significant distance range.
	      \emph{Disadvantage:} The 2D-LiDAR-based approach is highly sensitive to visibility conditions and performs poorly during poor weather conditions.
	\item
	      \textbf{GRAPH-SLAM} (\citeyear{holder2019_RealPose})~\cite{holder2019_RealPose} utilises a stochastic gradient descent approach for nonlinear optimisation.
	      GRAPH-SLAM uses radar sensors to perform point matching with ICP.
	      \emph{Advantage:} The approach uses the higher range and angular resolutions in radar for performing SLAM over long tracks.
	      \emph{Disadvantage:} GRAPH-SLAM can be sensitive in the choice of parameters and require fine-tuning.
	\item
	      \textbf{Particle Filter SLAM} (\citeyear{chen2020_ImprPart})~\cite{chen2020_ImprPart} uses Monte Carlo sequence filtering method for maintaining an estimated distribution of the current robot state.
	      \emph{Advantage:} The filtering process is performed with state identification, mass modification and a resampling procedure.
	      \emph{Disadvantage:} It requires lots of particles to perform state estimation in an environment with a large spatial area; otherwise, the likelihood will be spatially separated with large separation.
	\item
	      \textbf{Direct Sparse Mapping (DSM)} (\citeyear{zubizarreta2020_DireSpar})~\cite{zubizarreta2020_DireSpar} adopted the Photometric bundle adjustment (PBA) method for SLAM, which was shown to be effective for estimating scene geometry and camera motion in Visual Odometry (VO).
	      Unlike PBA, which estimates the camera odometry with a temporary map, DSM can build a persistent map for SLAM usage.
	      \emph{Advantage:} DSM is a direct monocular VSLAM method that detects point observations and extracts the geometric information from the photometric formulation.
	      \emph{Disadvantage:} PBA is needed during the DSM procedure, significantly increasing the runtime processing cost.
\end{itemize}

\begin{figure}[tb]
	\centering
	\includegraphics[width=0.85\linewidth]{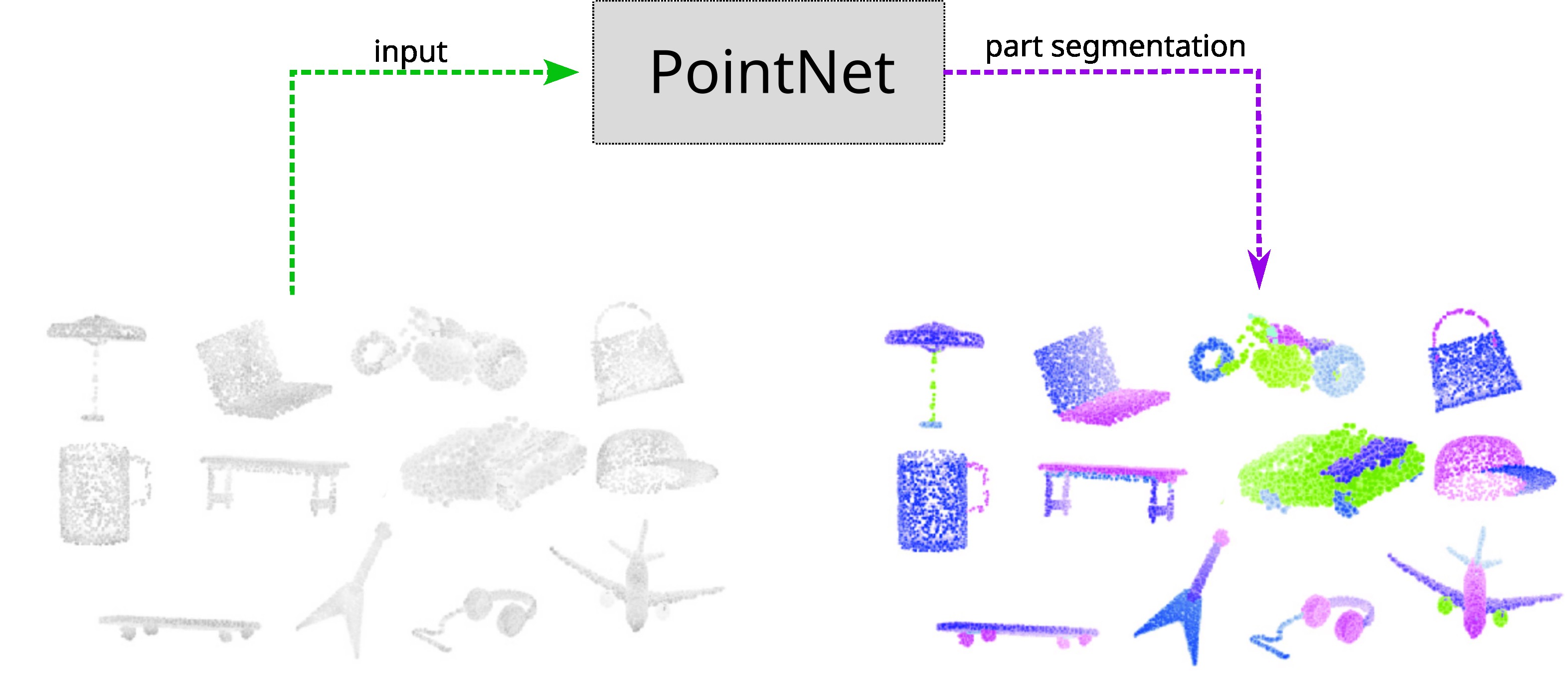}
	\caption{Example of using \textit{PointNet}~\cite{qi2017_PoinDeep} for performing part segmentation directly on input point clouds.
		\label{fig:scene-understanding-mono-modal}}
\end{figure}

\section{Visual-SLAM}

Visual-SLAM and sensors have been the main research direction for SLAM solutions due to their capability of collecting a large amount of information and measurement range for mapping.
The principle of Visual-SLAM lies in a sequential estimation of the camera motions depending on the perceived movements of pixels in the image sequence.
Besides robotics, Visual-SLAM is also essential for many enormous vision-based applications such as virtual and augmented reality.
Many existing Visual-SLAM methods explicitly model camera projections, motions, and environments based on visual geometry.
Recently, many methods have assigned and incorporated semantic meaning to the observed objects to provide a more successful localisation that is robust against observation noise and dynamic objects.
In this section, we will review the different families of algorithms within the branch of Visual-SLAM.

\subsection{Feature-based and Direct SLAM}

Feature-based SLAM can be divided into filter-based and Bundle Adjustment based methods introduced earlier in previous sections.
Earlier SLAM approaches utilised EKFs for estimating the robot pose while updating the landmarks observed by the robots simultaneously~\cite{guivant2001_OptiSimu, leonard2000_CompEffi, lu1997_GlobCons}.
However, the computational complexity of these methods increased with the number of landmarks, and they did not efficiently handle non-linearities in the measurements~\cite{bailey2006_ConsEKFS}.
FastSLAM was proposed to improve the EKF-SLAM by combining particle filters with EKFs for landmark estimation~\cite{montemerlo2002_FastFact}.
However, it also suffered from the limitations of sample degeneracy when sampling the proposal distribution.
Parallel Tracking and Mapping~\cite{klein2007_ParaTrac} was proposed to address the issue by splitting the pose and map estimation into separate threads, which enhance their real-time performance~\cite{castle2008_VideLoca, pradeep2013_MonoReal}.

A place recognition system with ORB features was first proposed in~\cite{mur-artal2014_FastRelo}, which is developed based on Bag-of-Words (BoW).
The ORB is a rotational invariant and scale-aware feature~\cite{rublee2011_ORBEffi}, which can be used to extract features at a high frequency.
Place recognition algorithms can often be highly efficient and run in real time.
The algorithm is helpful in relocalisation and loop-closure for Visual-SLAM, and it is further developed with monocular cameras for operating in a large-scale environment~\cite{endres2013_3DMapp}.

RGB-D SLAM~\cite{endres2013_3DMapp} is another feature-based SLAM that uses feature points for generating dense and accurate 3D maps.
Several models are proposed to utilise the active camera sensor to develop a 6-DOF motion tracking model capable of 3D reconstruction and achieve impressive performance even under challenging scenarios~\cite{kueng2016_LowlVisu,kim2016_Real3D}.
In contrast to low-level point features, high-level objects often provide a more accurate tracking performance.
For example, using a planar SLAM system, we can detect the planar in the environment for yielding a planar map while detecting objects such as desks and chairs for localisatino~\cite{salas-moreno2013_SlamSimu}.
The recognition of the objects, however, requires an offline supervised-learning procedure before executing the SLAM procedure.

\emph{Direct SLAM} refers to methods that directly use the input images without any feature detector and descriptors.
In contrast to feature-based methods, these feature-less approaches are generally used in photometric consistency to register two successive images.
Using deep-learning models for extracting the environment's feature representation is promising in numerous robotic domains~\cite{hewing2020_LearMode,lai2021_PlanLear}.
For example, DTAM~\cite{newcombe2011_DTAMDens}, LSD-SLAM~\cite{engel2014_LSDSLarg} and SVO ~\cite{forster2014_SVOFast} are some of the models that had gain lots of successes. DSO models~\cite{engel2017_DireSpar, wang2017_SterDSO} are also shown to be capable of using bundle adjustment pipeline of temporal multi-view stereo for achieving high accuracy in a real-time system.
In additions, models such as CodeSLAM~\cite{bloesch2018_CodeComp} and CNN-SLAM~\cite{tateno2017_CnnsReal} use deep-learning approach for extracting a dense representations of the environment for performing direct SLAM.
However, direct SLAM is often more time-consuming when compared to feature-based SLAM since they operate directly on the image space.

\begin{figure}[tb]
	\centering
	\includegraphics[width=0.75\textwidth]{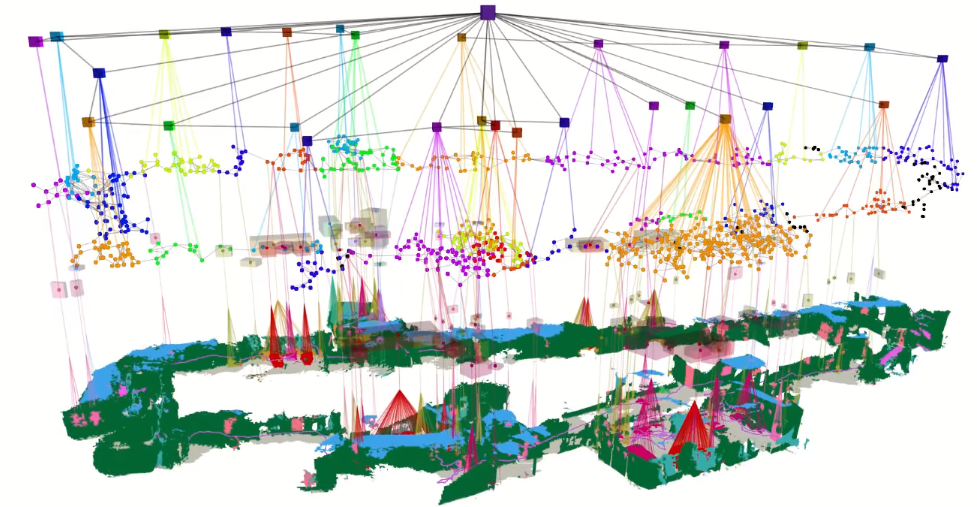}
	\caption{Using \textit{Dynamic Scene Graph (DSG)} \cite{rosinol2021_KimeSLAM} for generating multi-layer abstraction of an indoor environment.
		\label{fig:dsg-algorithm}}
\end{figure}

\subsection{Localisation with Scene Modelling}

Deep learning plays an essential role in scene understanding by utilising a range of information in techniques such as CNN classifications.
CNN can be utilised over RGB images for extracting semantic information like detecting scene or pedestrians within the images~\cite{long2015_FullConv,geraldes2019uav,peng2021_CrosFusi}.
CNN can also directly operates on captured point cloud information from range-based sensors like LiDAR.
Models like PointNet~\cite{qi2017_PoinDeep} in~\cref{fig:scene-understanding-mono-modal} can understand classifying the class of the objects based purely on point clouds.
For example, PointNet++~\cite{qi2017_PoinDeep}, TangentConvolutions~\cite{tatarchenko2018_TangConv}, DOPS~\cite{najibi2020_DopsLear}, RandLA-Net~\cite{hu2020_RandEffi} are some of the recent deep learning model that can perform semantic understanding using a large scale of point clouds.
Most models are trained on some point cloud dataset that enables the model to infer objects and scene information based purely on the geometric orientations of the input points.

\begin{figure}[tb]
	\centering
	\includegraphics[width=0.405\linewidth]{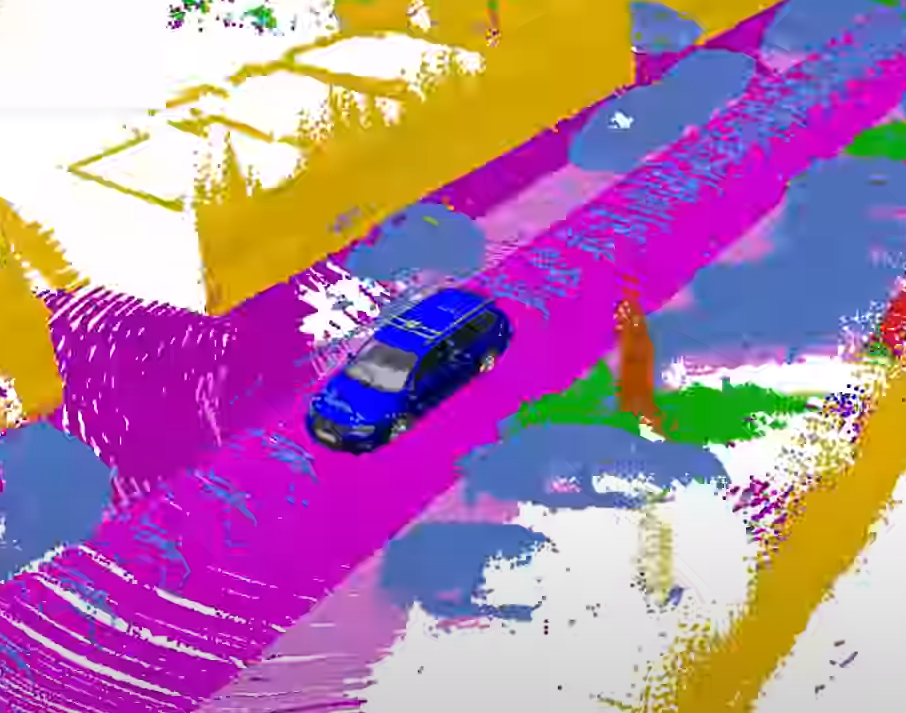}\hfill
	\includegraphics[width=0.55\linewidth]{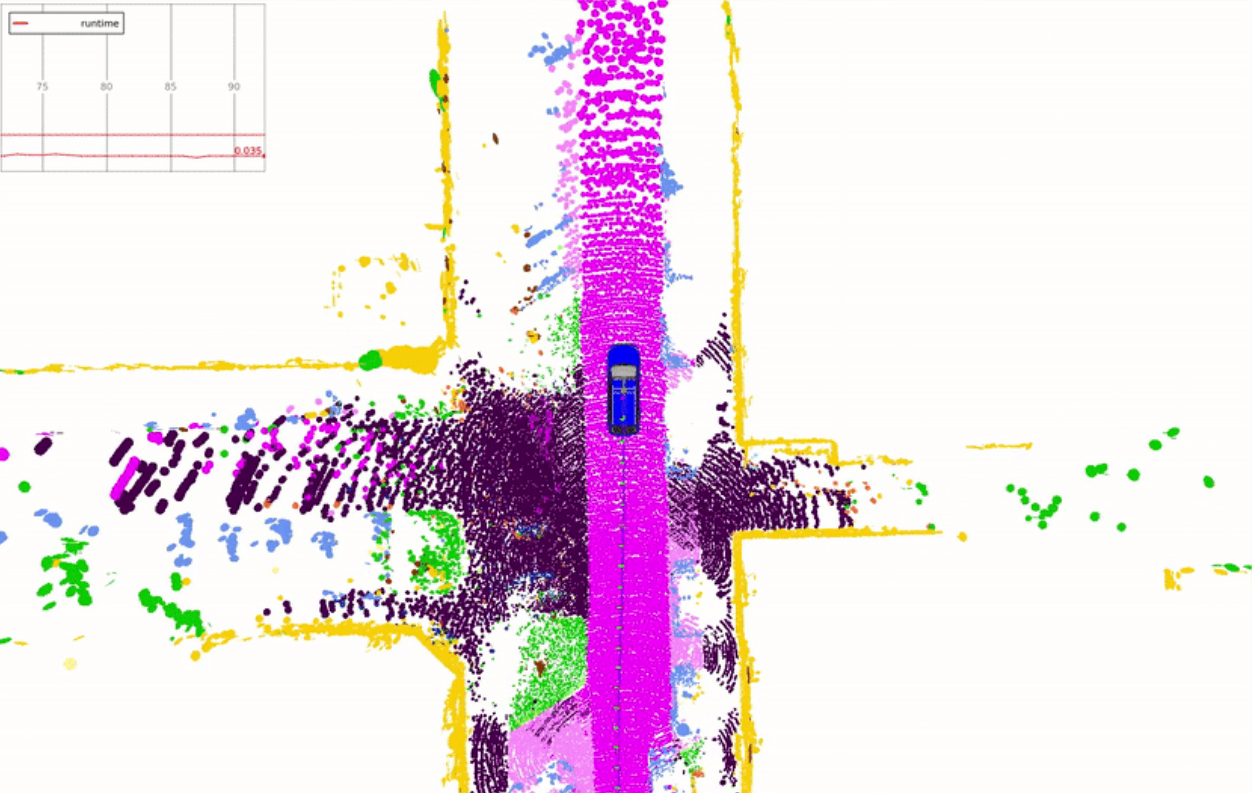}
	\caption{Using environment features to create a semantic map.
		\textit{SUMA++} \cite{chen2019_SumaEffi} operating under an environment using LiDAR sensors, which provides rich information to understand the environment around the vehicle.
		\label{fig:featur-map-understanding}}
\end{figure}

Dynamic objects can introduce difficulties in SLAM during loop-closure due to the moving objects.
SLAM can tackle this difficulty by utilising semantics information to filter dynamic objects from the input images~\cite{wang2019_CompEffi}.
Using the scene understanding module, we can filter out moving objects from the images to prevent the SLAM algorithm conditioning on dynamic objects.
For example, the \emph{SUMA++} model illustrated on the right of~\cref{fig:featur-map-understanding} can obtain a semantic understanding of each detected object to filter out dynamic objects such as pedestrians and other moving vehicles.
However, the increased SLAM accuracy comes with the cost of lowering the accuracy of the estimated robot pose due to the method neglecting parts of the perceived information.

\subsection{Scene Modelling with Typological Relationship and Dynamic Models}

\emph{Scene graphs} is a different approach to building a model of the environment that includes both the metric, semantic, and primary topological relationship between the scene objects and the overall environment~\cite{armeni2019_3dScen}.
Scene graphs can construct an environmental graph that spans an entire building, including objects, materials and rooms within the building~\cite{wald2020_Lear3d}.
The main disadvantage of scene graphs is the need to compute offline, requiring a known 3D mesh of the building with the registered RGB images to generate the 3D scene graphs.
Previous approaches rely on registering RGB images with the 3D mesh of the buildings for generating the 3D scene graphs, which limits their applicability to static environments.
\Cref{fig:dsg-algorithm} illustrates one of the approaches, Dynamic scene graphs (DSG)~\cite{rosinol2021_KimeSLAM}, that can also include dynamic elements within the environment.
For example, DSG can model humans that are navigating within the building.
The original DSG approach needs to be built offline, but an extension has been proposed~\cite{rosinol2021_KimeSLAM} which is capable of building a 3D dynamic DSG from visual-inertial data in a fully automatic manner.
The approach first builds a 3D mesh-based semantic map fed to the dynamic scene generator.

In addition, we can perform reasoning on the current situation by projecting what will likely happen based on previous events~\cite{castillo-lopez2020_RealAppr}.
This class of methods relies on predicting the possible future state of the robot by conditioning on the current belief of our robot state and the robot's dynamic model~\cite{sanchez-lopez2017_VisuMark}.
In addition, dynamic models can be incorporated into the objects in the surrounding environment, such as pedestrians and vehicles, for the model to recognise the predicted future pose of the nearby objects with some amount of uncertainty~\cite{lefkopoulos2020_InteMoti}.

\subsection{Semantic Understanding with Segmentation}

Pixel-wise semantic segmentation is another promising direction in SLAM semantic understanding.
FCN~\cite{long2015_FullConv} is a fully convolutional neural network that uses pixel-wise segmentation in the computer vision community for SLAM.
ParseNet derived a similar CNN architecture~\cite{liu2015_ParsLook} and injected the global context information into the global pooling layers in FCN.
The global context information allows the model to achieve better scene segmentation with a more feature-rich representation of the network.
SegNet is another novel netowkr~\cite{badrinarayanan2017_SegnDeep} that uses an encoder-decoder architecture for segmentation.
The decoder architecture helps upsample the captured low-resolution features from the images.
Bayesian approaches are helpful in many learning-based robotics application~\cite{kendall2015_BayeSegn,lai2020_BayeLoca}.
Bayesian SegNet~\cite{kendall2015_BayeSegn} took a probabilistic approach by using dropout layers in the original SegNet for sampling.
The Bayesian approach estimates the probability for pixel-level segmentation, which often outperforms the original approach.
Conditional Random Fields had been combined with CNN architecture~\cite{zheng2015_CondRand} for deriving a mean-field approximate inference as Recurrent neural Networks.

Semantic information is particularly valued in an environment where a robot needs to interact with human~\cite{hulse2010_FastLear}.
The progress in computer vision semantic segmentation using deep learning is constructive for pushing the research progress in semantic SLAM.
By combining model-based SLAM methods with spatio-temporal CNN-based semantic segmentation~\cite{li2018_SemaScen}, we can often provide the SLAM model with a more informative feature representation for localisation.
The proposed system can simultaneously perform 3D semantic scene mapping and 6-DOF localisation even in a large indoor environment.
Pixel-voxel netowk~\cite{zhao2018_DensRgbd} is another similar approach that uses CNN-like architecture for semantic mapping.
SemanticFusion~\cite{mccormac2017_SemaDens} integrates the CNN-based semantic segmentation with the dense SLAM technology ElasticFusion~\cite{whelan2015_ElasDens}, resulting in a model that produces a dense semantic map and performs well in an indoor environment.

\begin{figure}[tb]
	\centering
	\includegraphics[width=0.95\linewidth]{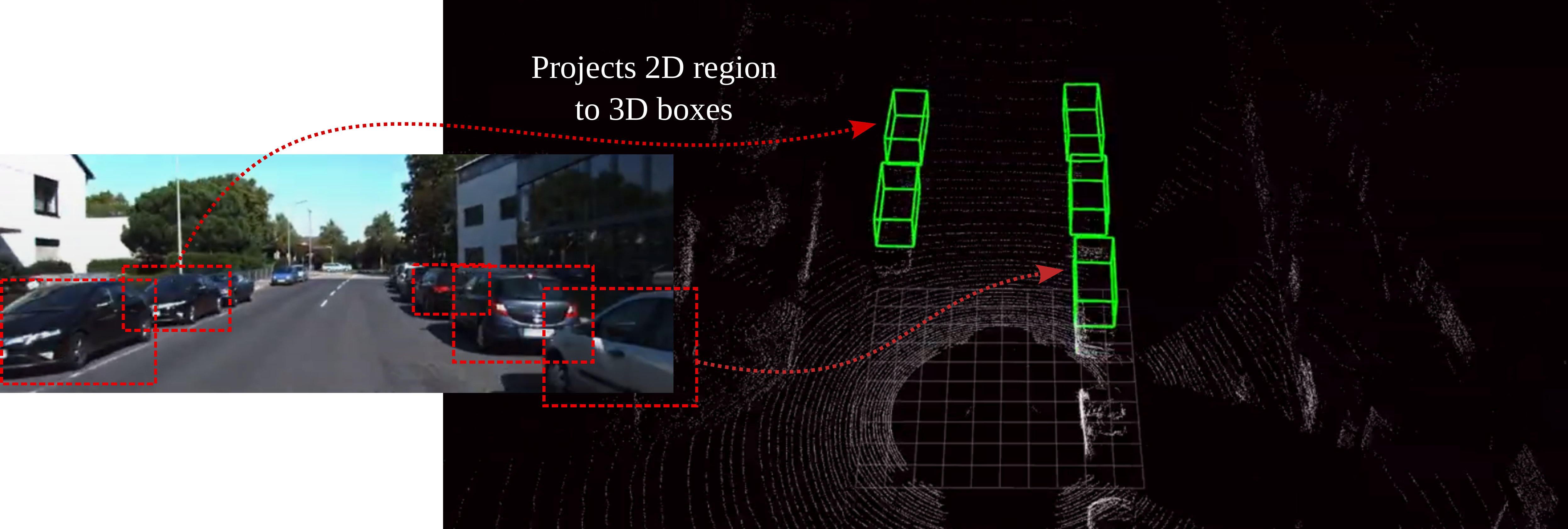}
	\caption{Multi-modal model \textit{Frustrum PointNets}~\cite{qi2018_FrusPoin} which uses CNN model to projects detected objects from \textit{RGB} images into 3D space, thus improving the accuracy on semantic understanding.
		\label{fig:scene-understanding-multi-modal}}
\end{figure}

\subsection{Sensors Fusions for Semantic Scene Understanding}

With the recent advancements in Deep Learning, numerous Visual-SLAM have also gained treatment success in using the learned models for semantic understanding using data fusion.
Models such as Frustrum PointNets~\cite{qi2018_FrusPoin} utilise both RGB camera and LiDAR sensors to improve the accuracy of understanding the semantics of the scene.
\Cref{fig:scene-understanding-multi-modal} illustrates how Frustrum PointNet utilises information from both sensors for data fusion, where a PointNet is first applied for object instance segmentation and amodal bounding box regression.
Sensor fusion provides a more rich feature representation for performing data association.
For example, VINet is a sensor fusion network~\cite{clark2017_VineVisu} that can use the estimated pose from DeepVO~\cite{wang2017_DeepEndt} along with the inertial sensor readings with an LSTM.
During the model training procedure, the prediction and the fusion network are trained jointly to allow the gradient to pass through the entire network.
Therefore, both networks can compensate each other, and the fusion system has high performance compared to traditional sensor fusion methods.
The same methodology can also be used as a fusion system~\cite{turan2017_EndoDeep} which is capable of fusing the 6-DOF pose data from the cameras and the magnetic sensors~\cite{turan2018_EndoPart}.

The information obtained from a camera can also be fused with GPS, INS, and wheel odometry readings as an ego-motion estimation system~\cite{pillai2017_VisuEgom}.
The model essentially uses deep learning to capture the temporal motion dynamics.
The motion from the camera is utilised in a mixture density network to construct an optical flow vector for better estimation.
Direct methods for visual odometry (VO) can often exploit information from the intensity level gathered from the input images.
However, these methods cannot guarantee optimality compared to feature-based methods.
Semi-direct VO (SVO2)~\cite{forster2017svo} is a hybrid method that uses direct methods to track pixels while relying on feature-based methods for joint optimisation of structure and motions.
The hybrid methods take advantage of both approaches to improve the robustness of VO.
Similar approaches like VINS-Fusion~\cite{qin2018vins} are capable of using IMU fused with monocular visual input for estimating odometry with high reliability.
Deep neural networks can further learn the rigid-body motion in a CNN architecture~\cite{byravan2017_Se3nLear} using raw point cloud data as input for predicting the SE3 rigid transformation of the robot.

\section{Conclusion and Future Directions}

Numerous studies have been conducted in the SLAM domain, as mapping and navigation are critical for enabling robots to interact autonomously with the real world.
SLAM algorithms remain a promising and exciting research domain due to their ubiquitous needs in mobile robotic applications.
SLAM merges ideas from multiple fields that bridge community within the broader robotic system, for example, sensing, perception, localisation and mapping.
In addition, a visual SLAM system with learning capability has shown tremendous potential for further exploration.
Approaches with deep learning are shown to be more flexible in producing a more robust approach via utilising the semantic information about the surrounding objects.
Sensors information such as pose, depth, 3D point cloud, and semantic mapping of the surrounding objects have shown to be highly useful in Visual-SLAM.
By fusing the measured readings from different sensors, the learning-based models can utilise more sources of information for a more feature-rich data-association process.
We believe a learning-based approach in semantic SLAM is a promising and exciting direction for developing autonomous robots.

SLAM provides the foundations for the autonomous operations of robots.
Many possible future directions can further improve the challenges discussed in earlier sections.
For example, data association is one of the core problems in SLAM.
Some current IMU and visual odometry-based approaches depend highly on sensors' accuracy or assume some prior on normally distributed and stationary noise.
Having an adaptive approach to tackling possible shifting temporal noise distribution can further mitigate the data association problem.
Sensor fusion should be another focus in Visual-SLAM due to the availability of various sensors in modern robots.
LiDAR and RGB-D cameras are the two most popular approaches in modern SLAM; therefore, combining the rich information provided by the sensors can further improve the current state-of-the-art Visual-SLAM algorithms.
Currently, the SLAM and motion planning problems in robotics are typically tackled in a disjointed manner.
However, integrating the uncertainty and probabilistic information obtained in the SLAM framework would theoretically provide more information for the robot to plan for its next movement during motion planning.
Therefore, integrating motion planning algorithms such as RRT or PRM within SLAM could provide a more robust robotic framework.
Finally, several works that depend on deep neural netowkrs have been discussed in previous sections.
Integrating methodologies in deep reinforcement learning literature can perhaps provide SLAM with a learnable policy that exploits past SLAM episodes to improve future execution in unseen environments.

We provide a thorough literature review of the fundamental and current state-of-the-art Visual-SLAM models for communicating our current understanding of SLAM approaches.
We have shown ongoing evolution in Visual-SLAM, from model-based approaches to deep learning-based methods.
Most current SLAM models seek to improve their accuracy and robustness in the high-level cognition and perception within the Visual-SLAM systems.
The key to most current semantic Visual-SLAM models lies in designing the network architecture, appropriate loss function, and the data representation of deep learning-based methods.
Therefore, the ongoing research progress in deep-learning models will further enhance the capability of Visual-SLAM models.

\printbibliography

\end{document}